\documentclass[11pt]{article}
\usepackage{acl2015}
\usepackage{times}
\usepackage{url}
\usepackage{latexsym}
\usepackage{graphicx}

\graphicspath{ {./images/} }

\title{Pagsusuri ng RNN-based Transfer Learning Technique sa Low-Resource Language}

\author{Dan John Velasco \\
  De La Salle University, Manila \\
  2401 Taft Ave, Malate, Manila, Philippines \\
  {\tt dan\_velasco@dlsu.edu.ph} \\}

\date{}

\begin{document}
\maketitle
\begin{abstract}
Ang mga low-resource languages tulad ng Filipino ay gipit sa accessible na datos kaya't mahirap gumawa ng mga applications sa wikang ito. Ang mga Transfer Learning (TL) techniques ay malaking tulong para sa low-resource setting o mga pagkakataong gipit sa datos. Sa mga nagdaang taon, nanaig ang mga transformer-based TL techniques pagdating sa low-resource tasks ngunit ito ay mataas na compute and memory requirements kaya nangangailangan ng mas mura pero epektibong alternatibo. Ang papel na ito ay may tatlong kontribusyon. Una, maglabas ng pre-trained AWD-LSTM language model sa wikang Filipino upang maging tuntungan sa pagbuo ng mga NLP applications sa wikang Filipino. Pangalawa, mag benchmark ng AWD-LSTM sa Hate Speech classification task at ipakita na kayang nitong makipagsabayan sa mga transformer-based models. Pangatlo, suriin ang performance ng AWD-LSTM sa low-resource setting gamit ang degradation test at ikumpara ito sa mga transformer-based models.
\end{abstract}

\section{Introduksyon}

Sa mga nagdaang taon, mabilis ang pag-unlad ng Natural Language Processing (NLP) dahil sa kasaganahan ng data dulot ng adopsyon ng internet sa buong mundo at sa pagiging accessible ng mas murang computing power. Nagkaron ng pag-unlad sa mga aplikasyon ng NLP tulad ng Machine Translation, Text Categorization, Text Classification, at iba pa.

Karamihan sa mga tagumpay ng NLP ay para sa mga mainstream na wika tulad ng Ingles at iba pang wika na merong accessible na malalaking text corpora at annotated texts. Ang mga wikang wala o limitado ang access sa malaking text corpora o annotated texts ay tinatawag na low-resource languages.

Ang pangangailangan ng malalaking text corpora o datasets para mapakinabangan ang mga benepisyo ng NLP ay isang balakid na pumipigil o nagpapabagal ng adopsyon ng teknolohiya sa mga low-resource languages. Sa papel na ito, pagtutuunan ng pansin ang paggamit ng Transfer Learning (TL) upang malagpasan ang balakid ng low-resource language tulad ng wikang Filipino. 

Sa mga nakaraang taon, nanaig ang mga transformer-based transfer learning techniques pagdating sa pagiging epektibo sa mga iba't ibang aplikasyon ng NLP sa low-resource languages tulad ng Fake News Detection in Filipino \cite{FAKENEWS}, Named Entity Recognition in Vietnamese \cite{PHOBERT}, at iba pa, 

Bagamat ang transformer-based models ay napatunayang epektibo sa iba't ibang aplikasyon ng NLP, ito ay hindi ganoon ka-accessible sa kasalukuyang panahon dahil mataas ang kailangang compute power at memory upang ma-train ito. Habang ang RNN-based models naman kagaya ng AWD-LSTM \cite{MERITY} ay mas higit na accessible o mura i-train kumpara sa transformer-based models. Ang accessibility ng AWD-LSTM ay mahalaga para mapalawak ang adopsyon ng NLP sa Pilipinas at mas mapakinabangan ng nakararami ang benepisyo nito.

Ang papel na ito ay may tatlong kontribusyon: 1) Maglabas ng pre-trained AWD-LSTM language model (LM) sa wikang Filipino upang maging tuntungan sa pagbuo ng mga NLP applications sa wikang Filipino\footnote{https://github.com/danjohnvelasco/Filipino-ULMFiT}. 2) Mag benchmark ng AWD-LSTM sa Hate Speech classification task at ipakita na kayang makipagsabayan nito sa mga transformer-based models. 3) Pangatlo, suriin ang performance ng AWD-LSTM sa low-resource setting gamit ang degradation test at ikumpara ito sa mga transformer-based models.

\section{Background}
\subsection{Natural Language Processing}
Ang Natural Language Processing (NLP) ay isang subfield ng linguistics, computer science, at artificial intelligence na nauukol sa pag proseso at pag-unawa ng natural na wika \cite{WIKI:NLP}. Ang ilan sa mga aplikasyon ng NLP ay ang email spam filters (Text Classification), pag-unawa ng nais sabihin tulad ng mga smart assistants (Language Understanding), pagsasalin ng isang wika sa iba pang wika (Machine Translation), mag predict ng susunod na salita base sa mga naunang salita (Language Modelling), at marami pang iba. Dahil sa kaunlaran sa kasaganahan sa datos at pagiging accessible ng malakas na compute power, nabuhay muli ang machine learning approach. Sa maikling salita, ang machine learning approach ay gumagamit ng malaking datos na ginagamit ng isang computer algorithm upang matutunan ang mga patterns ng datos na ito. Dahil dito, naging epektibo siyang approach sa mga komplikadong problema dahil hindi na kailangan direktang i-program ang mga rules para malutas ang isang problema.

\subsection{Transfer Learning}
Notorious ang machine learning approach sa pangangailangan nito ng sobrang laking datos para mapakinabangan. Ang Transfer Learning (TL) ay isang area ng research na concerned sa problemang ito \cite{WIKI:TL}. Sa maikling salita, ang TL ay ang pag retain o pagpapanatili ng mga natutunan ng isang model sa isang gawain at paggamit o "transfer" ng mga natutunan nito sa iba pero may kaugnayan na gawain. Halimbawa, ang mga natutunan ng isang model sa pag detect ng muka ng tao ay maaring gamitin bilang tuntungan para sa pag-aaral ng model na matutunan kung ang muka ng tao ay galit, masaya, at iba pang facial expressions \cite{LI}.

\section{Metodolohiya}
\begin{figure*}[t]
\includegraphics[width=16cm]{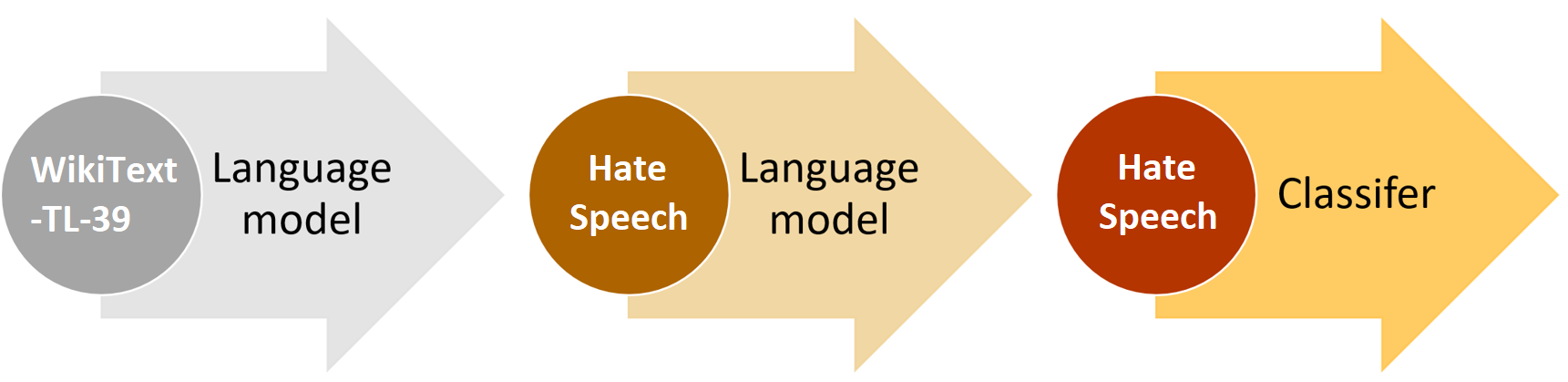}
\caption{ULMFiT approach summary. Adapted from fastai documentation\footnote{https://docs.fast.ai/tutorial.text}. Figure was modified to match the dataset used in this paper.}
\label{fig:ulmfit_summary}
\centering
\end{figure*}

Ang metodolohiya ay ang mga sumusunod 1) Mag train ng model gamit ang ULMFiT \cite{ULMFIT}, isang RNN-based transer learning method. 2) Sukatin ang kahusayan ng model sa text classification task gamit ang Hate Speech Dataset \cite{CABASAG}. Gamit ang parehong dataset, magsasagawa ng degradation test upang masuri ang performance ng AWD-LSTM sa low-resource setting.\cite{CRUZCHENG:20}. Makikita ang kabuuang proseso ng ULMFiT sa Figure \ref{fig:ulmfit_summary}.

Ang API na ginamit sa papel na ito ay fastai v2.0.13\footnote{https://pypi.org/project/fastai/2.0.13/}. Kapag hindi binanggit ang isang partikular na configuration sa model, ibig sabihin ang default setting lang ang ginamit. Para mas mapabilis ang training, gumamit ng mixed precision training \cite{MICIKEVICIUS}. Ang GPU na ginamit sa training ay Tesla T4. Sa buong proseso ng model training, ang learning rate schedule na 1cycle policy \cite{LESLIESMITH} ang ginamit. Ang buong code ay available sa public repository\footnote{https://github.com/danjohnvelasco/Filipino-ULMFiT}.

\subsection{ULMFiT}
Ang ULMFiT o Universal Language Model Fine-tuning \cite{ULMFIT} ay isang epektibong transfer learning technique na gumagamit ng language model na natuto sa malaking unlabeled text corpora at gagamitin ito bilang tuntungan sa iba pang gawain. Napatunayan na epektibo ang approach na ito kahit na maliit lang ang target text corpus o datos na gagamitin para sa partikular na gawain. Ang technique na ito ay may tatlong hakbang: 1) LM Pretraining Phase o pag pretrain ng AWD-LSTM language model sa isang malaking unlabeled text corpus. 2) LM Fine-tuning Phase o ang paggamit ng pre-trained LM bilang tuntungan at sanayin o i-train pa ng husto ang model sa target text corpus. 3) Text Classifier Fine-tuning Phase o ang pagsanay ng LM fine-tuned model sa text classification task.

\subsubsection{Language Model Pre-training}
Gumagamit ito ng AWD-LSTM \cite{MERITY} at sasanayin ito sa language modelling task na kung saan base sa isang pagkasunod sunod na mga salita, magbibigay ka ng prediksyon kung anong salita ang may pinakamataas na probability na kasunod nito. Sa LM pre-training phase, kailangan ng isang malaking unlabeled text corpora upang matutunan ng model ang mga patterns at structure ng wikang ginagamit sa text corpora. Ang ideal na text corpora ay dapat malaki, diverse, at nacacapture ang mga general properties ng isang wika. Ang ginamit na training data ay ang WikiText-TL-39 \cite{CRUZCHENG:19} na mula sa mga artikulo sa Tagalog Wikipedia \footnote{https://tl.wikipedia.org/wiki/Unang\_Pahina}. Pinagisa ang ang train, valid, and test set at randomly na kinuha ang 10\% ng data bilang validation set at ang natirang 90\% ay ginamit bilang training set. Ang text data ay dumaan sa preprocessing\footnote{Para sa partikular na preprocessing rules, bisitahin ang https://docs.fast.ai/text.core\#Preprocessing-rules} bago gamitin sa training. Kinukuha lamang dito ang 60,000 na salita na pinaka madalas makita sa datos. Ang model ay sinanay ng 20 epochs na may learning rate na 1e-2, batch size na 128, at dropout multiplier na 0.5. Ang buong proseso ng training ay tumagal ng 26 hours.

\subsubsection{Language Model Fine-tuning}
Gamit ang pre-trained model mula sa unang phase, mas sasanayin pa ang model sa target text corpus upang maka adapt ang model sa wika at kung ano mang patterns at vocabulary ang meron ito. Ang target text corpus na ginamit sa papel na ito ay ang Hate Speech Dataset \cite{CABASAG}. Ang dataset ay hinati sa train, validation, at test set. Sa pag fine-tune ng model, sinanay muna yung last layer ng model for 1 epoch na may learning rate na 4e-2. Pagkatapos nito, ang lahat naman ng layers ng model ay sasanayin for 7 epochs na may learning rate na 4e-3.

\subsubsection{Text Classifier Fine-tuning}
Gamit ang fine-tuned LM mula sa pangalawang phase, nag append ng karagdagang layers para sa text classification task \cite{ULMFIT}. Ang model ay muling sinanay sa target text corpus na Hate Speech dataset pero hindi na sa language modelling task kundi sa text classification task na. Dito, kasama na ang labels sa text (0 = not hate, 1 = hate). Ang dropout multiplier ay 0.3, weight decay ay 0.1, at momentum ay (0.8,0.7,0.6). Ginamit ang fine-tuning techniques na gradual unfreezing at discriminative learning rates \cite{ULMFIT}. Makikita sa Table \ref{tbl:finetune-table} ang buong proseso ng fine-tuning.

\begin{table}
  \includegraphics[width=\linewidth]{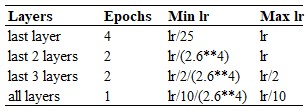}
  \caption{Set ng hyperparameters na ginamit sa fine-tuning na may gradual unfreezing and discriminative learning rates. Ang lr ay learning rate at lr = 5e-2.}
  \label{tbl:finetune-table}
\end{table}

\subsection{Degradation Test}
Ang degradation test \cite{CRUZCHENG:20} ay isang paraan ng pagsukat ng resillience ng model sa performance degradation kapag binawasan ang training samples. Ang performance degradation ay nirereport bilang percentage drop ng metric ng isang task. Ito ay: 

\[Degradation\% = \frac{Metric_{full}-Metric_{reduced}}{Metric_{full}}\]

na kung saan ang full ay ang performance sa kapag nag train sa buong training set at reduced ay ang performance kapag nag train sa reduced training set.

\begin{table*}[t]
\centering
\includegraphics[width=16cm]{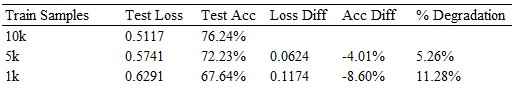}
\caption{Degradation test results sa Hate Speech Dataset}
\label{tbl:degradation-table}
\end{table*}

\begin{table}
  \includegraphics[width=\linewidth]{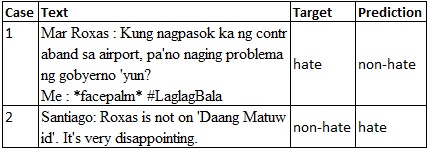}
  \caption{Halimbawa ng mga top losses ng model. Ang target ang tamang label at ang prediction ay ang prediction ng model.}
  \label{tbl:weakpoints}
\end{table}

Ang model na may mabagal na degradation ay mas epektibo sa low-resource setting. Gamit ang Hate Speech dataset, hahatiin sa ang proseso sa tatlong setup. Sa unang setup, gagamitin ang buong training set o 10k samples. Sa pangalawang setup ay 50\% split o 5k training samples nalang at ang panghuli ay 10\% split o 1k training samples. Ang buong proseso ng training sa bawat setup ay kapareho lang ng nabanggit sa Section 3. Limang beses itong uulitin sa bawat setup at kukunin ang average test loss at accuracy nito sa bawat setup. Makikita sa Table \ref{tbl:degradation-table} ang resulta ng degradation test.

\section{Resulta at Diskusyon}
\subsection{Resulta ng Fine-tuning}
Matapos ang fine-tuning sa Hate Speech dataset, ang AWD-LSTM ay naka score ng 76.84\% accuracy sa test set ng Hate Speech dataset. Ito ay lamang ng 2.08\% sa best model sa baseline \cite{CRUZCHENG:20} ngunit marginal improvement lamang ito.

\subsection{Resulta ng Degradation Test}
Ang AWD-LSTM ay may nagkaron ng performance drop sa accuracy na 4.01\% sa 5k split na may degradation na 5.26\%. At sa 1k split naman, sobrang laki ng binaba nito sa accuracy na bumaba ng 8.6\% at may degradation na 11.28\%.

Mula sa mga resultang ito, makikita na mas epektibo lang ng kaunti ang AWD-LSTM sa Hate Speech classification task kumpara sa BERT models kapag ginamit ang 10k training samples. Ngunit ang AWD-LSTM ay mas mababa ang performance pagdating sa mga pagkakataon na maliit lang ang datos tulad ng sa 5k at 1k training samples. Ang initial baseline ng BERT \cite{CRUZCHENG:20} ay may degradation na 3.28\% on average sa 5k split. Ang worst model sa baseline o ang pinakamabilis ang degradation ay ang DistilBERT \cite{DISTILBERT} na may degradation na 4.34\%  sa 5k split. Samantala ang AWD-LSTM naman ay may 5.26\% na degradation sa 5k split. Makikita dito na mas epektibo ang transformer-based models kagaya ng BERT at DistilBERT pagdating sa mga low-resource tasks.

Hindi ito nakakagulat dahil ang transformer-based models ay nakadisenyo talaga para makakuha ng mas malalim na patterns mula sa data. Ang advantage lang ng AWD-LSTM ay mas mabilis itong i-train lalo na kapag from scratch gagawin ang pre-trained language model at kaya itong gawin sa isang GPU lang hindi kagaya ng transformer-based models na kailangan gumamit ng Tensor Processing Unit (TPU).

\subsection{Weak Points ng Model}
Makikita sa Table \ref{tbl:weakpoints} ang halimbawa ng mga kaso ng top losses o ang mga prediksyon ng model na may mataas na confidence level pero mali pala. Ang paraan ng pagsusulat na ginamit sa case 1 ay high-context kung saan ang kahulugan nito ay nakadepende sa context ng paguusap at hindi sa literal na kahulugan ng mga salitang ginamit. Isa pang maaring dahilan ay ang paggamit ng *facepalm* o mga expression na hindi ganoon kadalas gamitin. Dahil hindi ito madalas gamitin, ibig sabihin hindi ito madalas makita sa dataset. At dahil hindi ito madalas makita sa dataset, hindi matututunan ng lubusan ng model ang patterns at relationships ng salita o expression na ito.

Sa case 2, ang salitang "disappointing" ang may pinaka malaking impluwensiya sa prediction ng model. Ang hinala ay dahil bihirang salita lang ito sa dataset at madalas gamitin ang salitang ito sa negatibong konteksto. Napatunayan ang hinala dahil meron lang tatlo na occurences sa training set ang salitang "disappointing" at lahat ito ay naka label as "hate". Ito ang dahilan kung bakit mataas ang association ng salitang "disappointing" sa label na "hate" at naging malakas ang impluwensiya nito sa pag predict sa isang sentence as hate.

In general, mahina ang performance ng model kapag ang mga salitang ginamit ay hindi common. Para matugunan ang problema na ito, kinakailangan ng mas malaki at mas diverse na dataset. Dahil kapag mas malaki at mas representative ng real world ang dataset, mataas ang chance na mataas ang general performance ng isang model.

\section{Kongklusyon}
Kahit na mas malalim ang patterns na kayang makuha ng transformer-based models, kaya parin makipagsabayan ng AWD-LSTM sa simpleng text classification task ngunit ang performance nito ay naghihingalo kapag ang training data ay maliit. Naipakita sa papel na ito na Nananaig padin ang transformer-based models kagaya ng BERT pagdating sa low-resource tasks. Nirerekomenda namin na gumamit ng transformer-based models kapag merong available na pre-trained language model sa iyong target na wika. At kapag from scratch gagawin ang pre-trained language model at kung issue ang oras at high compute power, mas mainam na gumamit ng AWD-LSTM dahil higit na mas mabilis ang pre-training step nito kumpara sa transformer-based models. Kung nais mong gumawa ng aplikasyon na kailangan gamitan ng NLP, nirerekomenda namin na magsimula muna sa mas murang option na AWD-LSTM at tignan kung sapat na ang performance nito para magawa ang isang task. Kapag hindi pa sapat ang performance nito, saka lamang subukan ang transformer-based models.


\begin{thebibliography}{}

\bibitem[\protect\citename{{Cabasag et al.}}2019]{CABASAG}
Neil Vicente Cabasag, Vicente Raphael Chan, Sean Christian Lim, Mark Edward Gonzales, and Charibeth Cheng.
\newblock 2019.
\newblock {\em Hate speech in philippine election-related tweets: Automatic detection and classification using natural language processing.} 
\newblock Philippine Computing Journal, XIV No. 1 August 2019.

\bibitem[\protect\citename{{Cruz and Cheng}}2019]{CRUZCHENG:19}
Jan Christian Blaise Cruz and Charibeth Cheng.
\newblock 2019.
\newblock {\em Evaluating language model finetuning techniques for low-resource languages.}
\newblock arXiv preprint arXiv:1907.00409.

\bibitem[\protect\citename{{Cruz et al.}}2019]{FAKENEWS}
Jan Christian Blaise Cruz, Julianne Agatha Tan, and Charibeth Cheng. 
\newblock 2019.
\newblock {\em Localization of Fake News Detection via Multitask Transfer Learning.}
\newblock arXiv preprint arXiv:1907.00409.

\bibitem[\protect\citename{{Cruz and Cheng}}2020]{CRUZCHENG:20}
Jan Christian Blaise Cruz and Charibeth Cheng.
\newblock 2020.
\newblock {\em Establishing Baselines for Text Classification in Low-Resource Languages.}
\newblock arXiv preprint arXiv:2005.02068.

\bibitem[\protect\citename{{Howard and Ruder}}2018]{ULMFIT}
Jeremy Howard and Sebastian Ruder.
\newblock 2018.
\newblock {\em Universal language model fine-tuning for text classification.}.
\newblock {In Proceedings of the 56th Annual Meeting of the Association for Computational Linguistics}, (Volume 1: Long Papers), pages 328–339.
\newblock Melbourne, Australia, July. Association for Computational Linguistics.

\bibitem[\protect\citename{{Li et al.}}2019]{LI}
Jianjun Li, Siming Huang, Xin Zhang, Xiaofeng Fu, Ching-Chun Chang, Zhuo Tang, Zhenxing Luo.
\newblock 2019.
\newblock {\em Facial Expression Recognition by Transfer Learning for Small Datasets.}
\newblock Part of the Advances in Intelligent Systems and Computing book series (AISC, volume 895)

\bibitem[\protect\citename{{Micikevicius et al.}}2017]{MICIKEVICIUS}
Paulius Micikevicius, Sharan Narang, Jonah Alben, Gregory Diamos, Erich Elsen, David Garcia, Boris Ginsburg, Michael Houston, Oleksii Kuchaiev, Ganesh Venkatesh, and Hao Wu.
\newblock 2017.
\newblock {\em Mixed Precision Training.}
\newblock arXiv preprint arXiv:1710.03740.

\bibitem[\protect\citename{{Merity et al.}}2017]{MERITY}
Stephen Merity, Nitish Shirish Keskar, and Richard Socher.
\newblock 2017.
\newblock {\em SRegularizing and Optimizing LSTM Language Models.}
\newblock arXiv preprint arXiv:1708.02182.

\bibitem[\protect\citename{{Nguyen and Nguyen}}2020]{PHOBERT}
Dat Quoc Nguyen and Anh Tuan Nguyen.
\newblock 2020.
\newblock {\em PhoBERT: Pre-trained language models for Vietnamese.}
\newblock arxiv preprint arXiv:2003.00744

\bibitem[\protect\citename{{Sanh et al.}}2019]{DISTILBERT}
Victor Sanh, Lysandre Debut, Julien Chaumond, and Thomas Wolf.
\newblock 2019.
\newblock {\em Distilbert, a distilled version of bert: smaller, faster, cheaper and lighter.}
\newblock arXiv preprint arXiv:1910.01108.

\bibitem[\protect\citename{{Smith and Topin}}2017]{LESLIESMITH}
Leslie N. Smith and Nicholay Topin.
\newblock 2017.
\newblock {\em Super-Convergence: Very Fast Training of Neural Networks Using Large Learning Rates.}
\newblock arXiv preprint arXiv:1708.07120.

\bibitem[\protect\citename{{Natural language processing}}n.d.]{WIKI:NLP}
Wikipedia.
\newblock n.d. {\em Natural language processing.}
\newblock Retrieved from https://en.wikipedia.org/wiki/
\newblock Natural\_language\_processing.

\bibitem[\protect\citename{{Transfer learning}}n.d.]{WIKI:TL}
Wikipedia.
\newblock n.d. {\em Transfer learning.}
\newblock Retrieved from https://en.wikipedia.org/wiki/Transfer\_learning

\end{thebibliography}
\end{document}